# Detecting Fake News on Social Media: A Novel Reliability Aware Machine-Crowd Hybrid Intelligence-Based Method

Yidong Chai[1], Kangwei Shi[1], Jiaheng Xie[2], Chunli Liu[1],
Yuanchun Jiang[1], Yezheng Liu[1]

[1]*School of Management, Hefei University of Technology*

[2]*Department of Accounting and MIS, University of Delaware*

*{chaiyd, liuchunli, ycjiang, liuyezheng}@hfut.edu.cn; shikw@mail.hfut.edu.cn; jxie@udel.edu*

## Abstract

Fake news on social media platforms poses a significant threat to societal systems, underscoring the urgent need for advanced detection methods. The existing detection methods can be divided into machine intelligence-based, crowd intelligence-based, and hybrid intelligence-based methods. Among them, hybrid intelligence-based methods achieve the best performance but fail to consider the reliability issue in detection. In light of this, we propose a novel Reliability Aware Hybrid Intelligence (RAHI) method for fake news detection. Our method comprises three integral modules. The first module employs a Bayesian deep learning model to capture the inherent reliability within machine intelligence. The second module uses an Item Response Theory (IRT)-based user response aggregation to account for the reliability in crowd intelligence. The third module introduces a new distribution fusion mechanism, which takes the distributions derived from both machine and crowd intelligence as input, and outputs a fused distribution that provides predictions along with the associated reliability. The experiments on the Weibo dataset demonstrate the advantages of our method. This study contributes to the research field with a novel RAHI-based method, and the code is shared at https://github.com/Kangwei-g/RAHI. This study has practical implications for three key stakeholders: internet users, online platform managers, and the government.

## 1. Introduction

Social media platforms like Twitter and Weibo empower users to share news from their own perspectives and experiences (Hu 2021, Hu and Hong 2022). With millions of users engaging daily, news on social media spreads rapidly (Fang et al. 2024), making it a significant channel for information dissemination (Kahr et al. 2024). According to a 2021 Pew Research Center survey, 48% of American adults frequently rely on social media for news. However, this surge in news sharing also comes with a proliferation of fake news (Ul Hussna et al. 2021). Worse still, fake news tends to spread much faster and wider than true news (García Lozano et al. 2020), posing significant risks to society and becoming a pressing social issue. For instance, during the COVID-19 pandemic, misinformation suggesting that disinfectants could treat the virus circulated widely and tragically led to three fatalities after consuming hand sanitizer in New Mexico. For another example, social bots manipulate and disseminate fake news, thereby distorting online discussions, influencing public opinion, and posing a significant threat to societal systems, such as the integrity of presidential elections (Mendoza et al. 2024). Such instances underscore the dire consequences of fake news on social media and the urgent need for advanced detection methods.

A plethora of methods have been proposed by both social media companies and academic researchers for the detection of fake news (Cardoso et al. 2019). Given the overwhelming volume of content on social media platforms, discerning what is true and what is not requires a sophisticated understanding of the information at hand. Intelligence plays a pivotal role in this process. These detection methods can be categorized into three types based on the intelligence they leverage (Collins et al. 2021). The first type relies on machine intelligence and is thus referred to as *machine intelligence-based methods* (Mundra et al. 2023). These methods employ various machine learning models to extract intelligence such as learning useful discriminative features from various sources like news publishers (Hayawi et al. 2023), text bodies (Chen et al. 2023), and images (Uppada et al. 2023) to assess the veracity of news articles. However, these methods are solely dependent on historical data, making them less effective when faced with emerging news stories (Hu et al. 2022). The second type of methods harness intelligence from a crowd of users and are therefore called *crowd intelligence-based methods* (Tchakounté et al. 2020). Users' responses to fake and truthful news articles differ due to their own knowledge and experiences (Li et al. 2022). For example, users may write refutations or comments to clarify when they perceive news as fake, while they may offer positive feedback if they believe the news is truthful (Souza

Freire et al. 2021). However, the reliability of crowd intelligence can be compromised by the presence of lay and uninformed users (Pennycook and Rand 2019), and the time required for user responses may hinder the early detection of fake news (Wu et al. 2023). The third type of methods combine both machine intelligence and crowd intelligence, termed as *hybrid intelligence-based methods* (Shabani and Sokhn 2018). Human crowds excel in deciphering the context and nuances of fake news (Jennifer Wortman Vaughan 2018), while machines supplement human with efficiency and the ability to detect news with few user responses (Liu and Wu 2020). By leveraging the strengths of both types of intelligence, hybrid intelligence-based methods achieve state-of-the-art performance in fake news detection (Qi et al. 2023, Shabani and Sokhn 2018, Wei et al. 2022).

However, hybrid intelligence-based methods overlook a critical aspect in detecting fake news: reliability. Unlike predictive probability which indicates the likelihood of a decision (e.g., being a certain class), reliability reflects the confidence and uncertainty of the decision. Higher reliability means the model is confident and more likely correct, while lower reliability suggests the situation may exceed the method's capacity, making the decision less trustworthy even if correct. Since news plays a vital role in informing the public and holding governments accountable, labeling news as fake requires caution. Thus, methods must assess both veracity and reliability. However, current hybrid intelligence methods neglect reliability in two ways. First, they fail to model reliability in machine intelligence. Studies (Yarin and Ghahramani 2016) show machine learning models inherently possess reliability, influenced by inadequacies and assumptions in modeling. For fake news detection, imperfect structures and parameters contribute to reliability issues. Second, while some studies model reliability in crowd intelligence by user reliability, limited social media engagement and variability among users compromise these measures. Efforts like Yang et al. (2019), which assess reliability based on historical user performance, improve accuracy but overlook varying news difficulty, further impacting crowd intelligence effectiveness.

The limitations of current hybrid intelligence-based methods motivate us to raise three questions in this study: 1) How can we model the reliability of machine intelligence in fake news detection? 2) How should we account for the difficulty of tasks in measuring user reliability when modeling the reliability of crowd intelligence? and 3) How can we effectively combine reliability in machine intelligence and uncertainty in crowd intelligence for fake news detection?

In response to the aforementioned research questions, this study proposes a novel Reliability Aware Hybrid Intelligence (RAHI) method for fake news detection, comprising three key modules. The first module uses a Bayesian deep learning model to capture inherent reliability in machine intelligence. Unlike traditional models with fixed parameters, Bayesian models treat parameters as distributions. Significant variability in the learned parameter distribution, such as a large standard deviation, indicates high unreliability. Assuming Gaussian-distributed assessment results, Bayesian deep learning provides predictions accounting for machine intelligence reliability. The second module employs Item Response Theory (IRT)-based user response aggregation to address reliability in crowd intelligence. IRT posits that item difficulty reflects user abilities. We compute news difficulty by dividing the number of users accurately identifying the news by the total participants. A monotonically increasing function models the relationship between difficulty and user reliability, calculated as the sum of derivatives of engaged item difficulties divided by the total items users engaged with. Since assessment results range from 0 (truthful) to 1 (fake), a Beta distribution models human assessments, offering predictions that incorporate crowd intelligence reliability. The third module is a distribution fusion mechanism combining machine and crowd intelligence assessments. To address challenges in fusing Gaussian and Beta distributions, we propose a likelihood maximization-based fusion mechanism (LMFM). Using a Multi-Layer Perceptron (MLP), we merge predictive distributions to generate a fused distribution, integrating the reliability of machine and human intelligence. The mean of the fused distribution determines fake news labels on social media.

We evaluate the advantages of the proposed RAHI method based on Weibo dataset. Four sets of experiments are conducted. In experiment 1, we conduct an overall performance analysis, assessing the extent to which the RAHI model outperforms other fake news detection methods. The results demonstrate that our RAHI method outperforms baseline methods. In experiment 2, we conduct ablation experiments to test the contribution of different components to the model's superior performance. The results demonstrate that each key design (i.e., considering reliability, combining human-crowd intelligence, updating user reliability) contributes to improved performance. In Experiment 3, we evaluate the model under scenarios of gradually increasing user responses. The results show that our method can operate well in dynamic scenarios where user response increase gradually. In Experiment 4, we conduct a case study to further demonstrate the details and the advantages of our method in fake news detection.

The contribution of this study is threefold. First, we introduce a reliability aware approach to detect fake news. By considering the reliability from machine intelligence and crowd intelligence, our method achieves superior accuracy in discerning the veracity of news. The awareness of reliability empowers practitioners to exercise greater prudence in detection. We have also shared our code on https://github.com/Kangwei-g/RAHI to facilitate future studies on fake news detection. Second, we propose a novel method for evaluating the user's reliability by considering the varying difficulty levels of different tasks. This method enables a more effective aggregation of user responses and facilitates a more accurate computation of crowd intelligence reliability. Our method holds promise for enhancing various research domains, particularly crowdsourcing, where the estimation of individual reliability is crucial for aggregating feedback. Third, we propose a novel method for integrating the Gaussian and the Beta distribution. While both Gaussian and Beta distributions are prevalent in research, the fusion of these two distributions remains unsolved. Leveraging neural networks, we devise a fusion method to attain a distribution that is close to both Gaussian and Beta distributions. Our approach not only facilitates the fusion of Gaussian and Beta distributions but also extends its applicability to fuse diverse forms of distributions with continuous output such as uniform distribution, exponential distribution, Gamma distribution, among others.

## 2. Literature Review: Fake News Detection in Social Media

Fake news refers to intentionally fabricated news articles designed to deceive readers (Dennis et al. 2023, Shu et al. 2017, Yuan et al. 2021). The content of fake news mimics truthful news but lacks the rigorous journalistic standards necessary for credibility (Mehta et al. 2022). Social networks offer convenient channels for disseminating information, making them fertile ground for the spread of fake news (Mocquard et al. 2020). Given the scale and seriousness of fake news on social media, various automated detection methods have been proposed to address this issue. According to the types of intelligence they exploited, the existing methods can be categorized into three types: machine intelligence-based detection, crowd intelligence-based detection and hybrid intelligence-based detection (Collins et al. 2021). The major studies of each type are summarized in Table A.1, Table A.2 and Table A.3 in Appendix A, and we will delve into each type next.

### 2.1 Machine Intelligence-Based Detection Methods

Machine intelligence-based detection methods utilize machine learning to identify fake news by learning distinguishing features from news content and social context. Fake news and truthful news exhibit different patterns in content, such as headlines, body text, images, and videos (Shu et al. 2017), as well as in social context, including user engagements and behaviors on social media. Early research employed traditional machine learning models like Logistic Regression (LR),

Support Vector Machine (SVM), Naive Bayes (NB), and K-Nearest Neighbor (KNN). For instance, (Granik and Mesyura 2017) used Naive Bayes to classify fake news based on linguistic features in news content. Kesarwani et al. (2020) combined user features from profiles with news features to detect fake news using KNN. With advancements in deep learning, neural networks have become the state-of-the-art for fake news detection, enabling the learning of complex features from content and social context (Alnabhan and Branco 2024). Wu et al. (2021) used a convolutional neural network (CNN) to extract textual features and VGG-19 to extract visual features from news, applying an attention mechanism to integrate these features for veracity assessment. Wu et al. (2023) leveraged BERT for extracting textual features from news and comments and VGG-19 for visual features, evaluating the coherence between comments, text, and images to detect fake news.

While valuable, the existing machine intelligence-based detection methods still have two limitations. First, they fail to consider the reliability behind veracity assessment. As proved in previous studies, machine learning models are inherent with uncertainty (Ovadia et al. 2019). Uncertainty reflects the reliability of the machine while it makes a decision. A lower uncertainty means the model is reliable in the current decision while a higher uncertainty suggests a less reliable decision. Though predictive prability can be obtained to indicate the likelihood of different classes, predictive probability can be misleading when it is used for measuring reliability (Yarin and Ghahramani 2016). Failing to consider the uncertainty makes the existing studies fail to consider the reliability of the prediction. This is particularly serious in fake news detection given the important role of news in our society. Hence, there is a pressing need for machine learning methods to consider the reliability in gauging the veracity of the news. The second limitation is that this type of methods purely relies on machine intelligence while overlooking another kind of useful intelligence, i.e., crowd intelligence.

### 2.2 Crowd Intelligence-Based Detection Methods

Crowd intelligence-based detection methods rely on users' responses to gauge the veracity of news, leveraging human judgment and thoughts. On social media, where users feel less restrained and express themselves more openly (Suler 2004), some users leave comments to alert others to fake news (Benjamin and Raghu 2023). Large groups are often collectively smarter than individuals in assessing news veracity (Souza Freire et al. 2021), making user responses valuable for fake news detection. Many platforms utilize responses such as comments, reposts, likes/dislikes, and reports. For example, Reddit allows users to cast votes, with a lower Reddit score reducing the spread of low-quality news. On Weibo, a major Chinese social media platform, users can submit suspicious comments or reports, which are reviewed by experts to verify news accuracy. A key challenge of crowd intelligence is the unreliability of user responses, as users may lack relevant knowledge or provide inappropriate judgments. Experienced users offer more reliable responses than inexperienced ones, making user reliability central to crowd intelligence. Current methods assess reliability using profiles and historical data. For instance, Yang et al. (2019) measure users' ability to correctly classify fake and true news based on historical behavior. Truong and Tran (2023) incorporate additional indicators, including the date joined, ratios of followers to followings, favorites to tweets, comments to tweets, retweets to tweets, and published fake news to tweets.

However, the existing crowd intelligence-based detection methods still have three limitations. First, they fail to consider the varying difficulty of gauging different pieces of news, and as a result, the obtained users' reliability is compromised. Particularly, a user should be more reliable if he or she can correctly gauge the veracity that fools most people. On the contrary, less reliability should be given to the user if the news has already been correctly identified by most people. The previous methods assume the difficulty of each piece of news is the same, and hence cannot effectively infer users' reliability. As a result, the obtained crowd intelligence and corresponding reliability are compromised. Second, the users' reliability is inferred separately from the task of fake news detection, and hence the direct application of the inferred user reliability on the downstream task may bring about suboptimal results. Third, they utilize crowd intelligence and fail to exploit machine intelligence, in contrast to machine intelligence-based methods.

### 2.3 Hybrid Intelligence-Based Detection Methods

Hybrid intelligence-based detection methods combine machine intelligence and crowd intelligence, leveraging their complementary strengths (Fügener et al. 2021). Human crowds, with their background knowledge, better understand news context and nuances and can seek external sources to verify suspicious news. In contrast, machine learning models are more efficient and can detect fake news early, even with minimal user response. By combining these strengths, hybrid methods achieve state-of-the-art performance in fake news detection. A critical component of hybrid methods is the fusion of machine and crowd intelligence. Current approaches include feature concatenation-based methods (Albahar 2021) and Bayesian network-based methods (Wei et al. 2022, Yang et al. 2019). Feature concatenation combines features from both intelligences into a unified set for detection (Albahar 2021). Bayesian network methods infer the veracity of news, a hidden variable, using Bayesian theorem, assuming that both machine responses (e.g., predictive probabilities) and crowd responses (e.g., the ratio of users identifying news as fake) are conditioned on the news's veracity (Wei et al. 2022, Yang et al. 2019).

However, the hybrid intelligence-based detection methods also inherit the disadvantages of both machine intelligence-based methods and crowd intelligence-based methods, i.e., the failure to consider reliability in machine intelligence, and the failure to consider the varying difficulty of gauging the veracity of different news in inferring users' reliability. As a result, the hybrid intelligence-based detection methods are still limited in providing reliability, which is critical for fake news detection. What is worse, it is even more challenging to provide reliability for hybrid intelligence-based detection methods because hybrid intelligence-based detection methods need to fuse the reliability from both machine intelligence and crowd intelligence. The existing methods mainly focus on fusing point results rather than fusing distributions. Fusing distributions captures the reliability and hence is perceived to be more effective, but it is also much more difficult. This necessitates a fusion method to fuse the distributions from both machine and crowd intelligence for fake news detection.

The advantages of hybrid intelligence-based methods motivate us to focus on this type in our study, while the existing limitations motivate us to raise the following three questions in this study: 1) How can we model the reliability of machine intelligence in fake news detection? 2) How should we account for the difficulty of tasks in measuring user reliability when modeling the reliability of crowd intelligence? and 3) How can we effectively fuse distributions from machine intelligence and uncertainty in crowd intelligence for fake news detection?

## 3. The Proposed Method: A Reliability-Aware Hybrid Intelligence-Based Method

In response to the three research questions, we propose a novel reliability-aware hybrid intelligence-based fake news detection method with three modules. We adopt deep learning to learn machine intelligence in our method. Hence, the first module in our proposed method models the reliability of prediction from a deep learning-based fake news detector. It outputs a Gaussian distribution which reflects the reliability of the predictions. The second module is to model the reliability of prediction from a crowd of users whose reliabilities are inferred by considering the difficulties of different tasks. It outputs a Beta distribution that reflects the reliability of the crowd's predictions. The third module fuses the Gaussian distribution and the Beta distribution by optimizing a certain distribution with maximum likelihood. The framework is shown in Figure 1.

**Figure 1. The Reliability-Aware Hybrid Intelligence Framework for Fake News Detection**

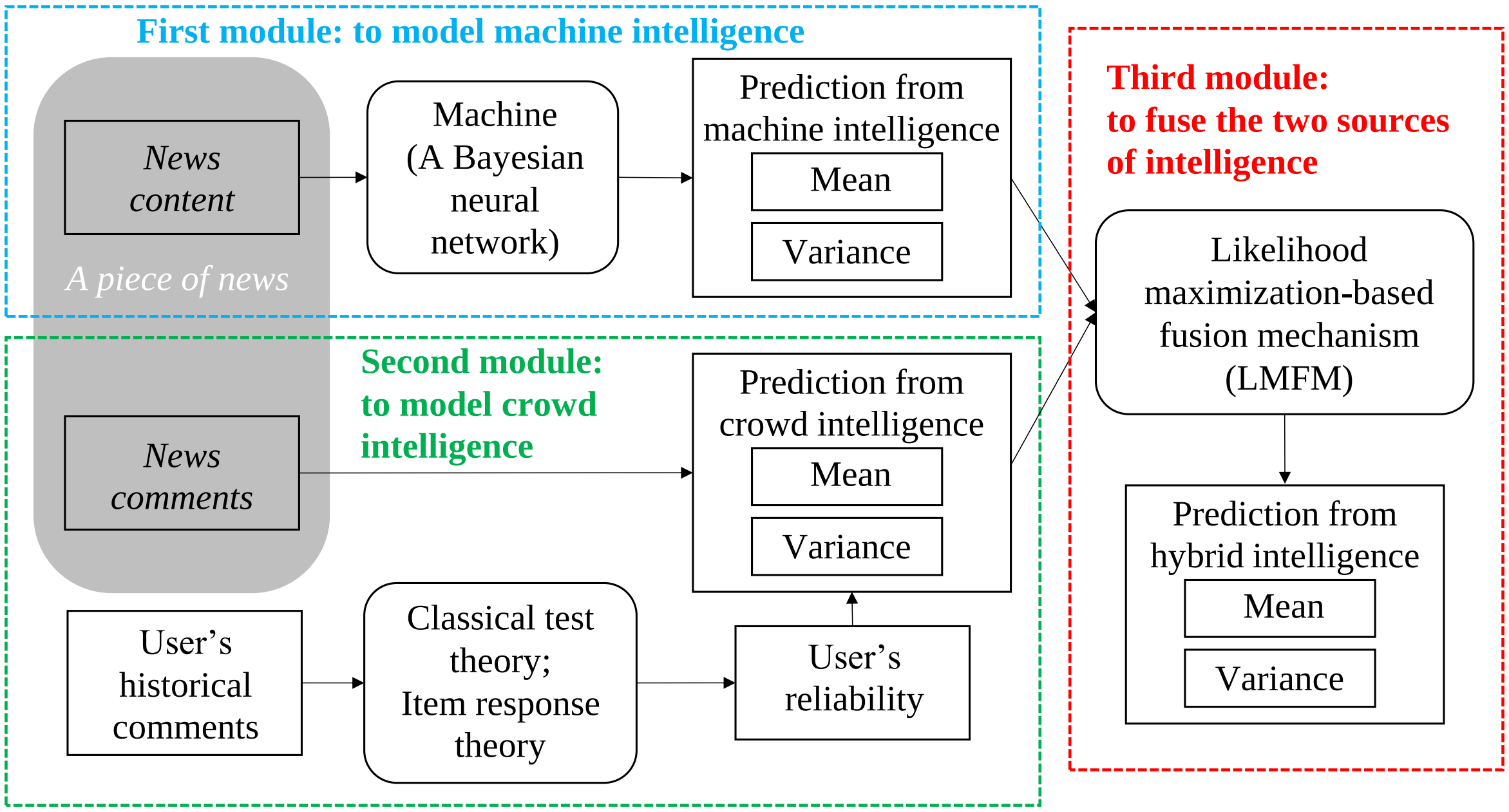

### 3.1 First Module: The Module to Model Machine Intelligence

As aforementioned, few fake news detection studies have considered the uncertainty in machine intelligence, thus failing to consider the reliability of the detection. However, the uncertainty in machine intelligence and deep learning in particular has become an important focus in areas such as healthcare, cybersecurity, and automated driving (Abdullah et al. 2022, Suk et al. 2024, Yang et al. 2024). Various methods have been proposed to model uncertainty, including ensemble learning, fuzzy theory, and Dempster-Shafer evidence theory-based methods. Among these methods, Bayesian deep learning stands out, offering clear insights into modeling uncertainty and providing a solid theoretical foundation (Gawlikowski et al. 2023). Therefore, it has become a major tool to model the machine intelligence uncertainty to inform the corresponding reliability. Hence, we also adopt Bayesian deep learning in this study.

Bayesian deep learning models the uncertainty of deep learning models in a Bayesian manner. Particularly, Bayesian deep learning incorporates the Bayesian theorem into deep learning where the model parameters follow a distribution and are updated based on the Bayesian theorem. Bayesian deep learning first puts prior distributions (denoted as $p(\boldsymbol{w})$) over the model's parameters (denoted as $\boldsymbol{w}$). Then, given observed samples $X = \{x_1, \dots, x_N\}$ and their corresponding labels $Y = \{y_1, \dots, y_N\}$, the posterior distribution (denoted as $p(\boldsymbol{w}|X, Y)$) is inferred based on the Bayesian theorem $p(\boldsymbol{w}|X, Y) = \frac{p(Y|X, \boldsymbol{w})p(\boldsymbol{w})}{p(Y|X)}$. With the learned parameters distributions ($p(\boldsymbol{w}|X,Y)$), the prediction made by the model is also a distribution given by $p(y|x) = \int p(y|x,\boldsymbol{w})p(\boldsymbol{w}|X,Y)\,d\boldsymbol{w}$. The distribution is obtained by considering all possible uncertain situations (i.e., the value of $\boldsymbol{w}$), and the final prediction is the expectation of the distribution.

A major challenge for Bayesian deep learning is the prohibitive inference process caused by the complex architecture of deep learning models. To address this challenge, various methods have been proposed, such as stochastic search (Paisley et al. 2012), variational Bayes (Kingma and Welling 2014), and probabilistic backpropagation (Miguel Hernández-Lobato and Adams 2015). Among them, Monte Carlo dropout (*MC dropout*) stands out as a convenient, widely applicable, and mathematically grounded method to approximate the posterior distribution (Gal 2016, Mukhoti and Gal 2018). MC dropout has been widely adopted in various tasks such as medical image classification and hydrological predictions. Hence, this study also adopts MC dropout for Bayesian deep learning in this study.

MC dropout directly applies a dropout operation before every neural network layer and the dropout operation remains active during the testing phase (Gal and Ghahramani 2016, Gawlikowski et al. 2023). Formally, denoting the weight of the neural network as $\boldsymbol{w}$. Then, in each feedforward process, we obtain a sample of weights, denoted as $\boldsymbol{m}$, whose value of each layer is obtained by randomly dropping the elements of $\boldsymbol{w}$. Though simple, it has been mathematically proven that the obtained $\boldsymbol{m}$ is actually sampled from the posterior distribution of $p(\boldsymbol{w}|X,Y)$ (Gal 2016). Following prior studies (Gawlikowski et al. 2023), we assume that the distribution of $p(y|x) = \int p(y|x,\boldsymbol{w})p(\boldsymbol{w}|X,Y)\,d\boldsymbol{w}$ is Gaussian $\mathcal{N}\left(E_{\text{machine}}(y|x), V_{\text{machine}}\,(y|x)\right)$, whose expectation is

$$E_{\text{machine}}(y|x) = \frac{1}{N}\sum_{n=1}^{N} p(y|x,\boldsymbol{m}_n) \tag{1}$$

where the $\boldsymbol{m}_n$ refers to the sample from the $n$-th feedforward process and $N$ is the number of feedforward processes. Meanwhile, the variance of distribution is given by

$$V_{\text{machine}}(y|x) = \frac{1}{N}\sum_{n=1}^{N}\left(p(y|x,\boldsymbol{m}_n) - \frac{1}{N}\sum_{n}^{N} p(y|x,\boldsymbol{m}_n)\right)^2 \tag{2}$$

The Gaussian distribution $\mathcal{N}\left(E_{\text{machine}}(y|x), V_{\text{machine}}\,(y|x)\right)$ can reflect the reliability of the decision made by machine intelligence.

### 3.2 Second Module: The Module to Model Crowd Intelligence

#### 3.2.1 Difficulty-Based User Reliability Estimation Method

Our estimation method is based on the Classical Test Theory (CTT) which posits that errors exist in the results of project testing and gives a simple evaluation method for the difficulty of the project. In our case, each fake news will attain many comments from different users, and the fake news is more difficult if only a smaller number of users respond correctly. We denote the difficulty of each news $i$ as $d_i$ and then $d_i$ is computed as

$$d_i = \frac{|U_i^t|}{|U_i|} \tag{3}$$

where $|U_i|$ denotes the number of users who leave comments on news $i$, and $|U_i^t|$ is the number of users who correctly identify news $i$ according to their comments.

According to Item Response Theory (IRT) which argues that there is a positive relationship between project difficulty and user ability. In our case, when a user can correctly gauge the veracity of a piece of news to which most users had incorrectly responded, this user is given more credit for their reliability. Then, we utilize the news difficulty $d_i$ to infer the reliability of a user $j$, which is denoted as $c_j$. Let $N_j$ denote the number of news items on which the user $j$ has left comments, and $C_j$ represent the set of news items that user $j$ correctly identifies. The reliability of user $j$ is then computed as

$$c_j = \frac{\sum_{i \in C_j} \frac{1}{d_i}}{N_j} \tag{4}$$

#### 3.2.2 Supervision-Guided User Reliability Adjustment

Based on two well-established theories (i.e., Classical Test Theory and Item Response Theory), $c_j$ can accurately reflect the reliability of each user. Then, the $c_j$ is utilized to aggregate the users' responses. For news $i$, we denote the set of users who believe the news is fake as $F_i$, while the users that believe the news is true as $T_i$. Then, the aggregated prediction that the news is fake is computed as

$$E_{\text{crowd}} = \frac{\sum_{j \in F_i} c_j}{\sum_{j \in F_i} c_j + \sum_{j \in T_i} c_j} \tag{5}$$

Here we introduce a supervision signal from the ground-truth to adjust the value of $c_j$, thereby further enhancing the accuracy of $c_j$ in reflecting the user's reliability. Hence, we call this adjustment method as supervision-guided user reliability adjustment. The predicted loss of crowd intelligence is given by,

$$L_{\text{crowd}} = \frac{1}{n_{\text{train}}} f_{\text{ce}}(E_{\text{crowd}}, y) = \frac{1}{n_{\text{train}}} \sum_{i=1}^{n_{\text{train}}} -[y \cdot \log E_{\text{crowd}} + (1-y) \cdot \log(1 - E_{\text{crowd}})] \quad (6)$$

where $n_{\text{train}}$ is the number of samples in the training data, $f_{\text{ce}}(E_{\text{crowd}}, y)$ is the cross-entropy function. Gradient descent-based optimization methods, such as SGD, are employed to learn $c_j$:

$$\boldsymbol{c} \leftarrow \text{SGDOptimizer}(\boldsymbol{\nabla} L_{\text{crowd}}, \boldsymbol{c}) \quad (7)$$

where $\boldsymbol{c}$ denotes the collection of $c_j$ for each user.

Since the response from users (i.e., $E_{\text{crowd}}$) is a binary variable, we assume it follows a Beta distribution $\text{Beta}(\alpha, \beta)$, where $\alpha$ and $\beta$ are two parameters of the distribution. $\alpha$ and $\beta$ are given by Equation (8) and (9):

$$\alpha = E_{\text{crowd}} \cdot |U_i| \quad (8)$$

$$\beta = (1 - E_{\text{crowd}}) \cdot |U_i| \quad (9)$$

The learned Beta distribution can reflect the reliability of the prediction.

### 3.3 Module of Fusion: A New Likelihood Maximization-Based Fusion Mechanism (LMFM)

Since the prediction from machine intelligence is a Gaussian distribution while the prediction from crowd intelligence is a Beta distribution, the fusion module outputs a distribution that contains the richest information from both sources. Fusing multiple distributions into one is common in the research field. Most previous studies fuse multiple homogenous distributions by minimizing the sum of the discrepancy between the existing distributions and the fused distribution. Formally, assuming there are $K$ distributions $\{f_k\}$, they aim to find a distribution $f_{\text{fused}}$ by

$$f_{\text{fused}} = \text{argmin}_f \sum_{k}^{K} DIS(f, f_k) \quad (10)$$

where $DIS(f, f_k)$ refers to the discrepancy between distribution $f$ and the distribution $f_k$. This approach operates well in the situation of fusing homogenous distributions because it is relatively easy to compute the discrepancy between two homogenous distributions such as two Gaussian distributions or two Beta distributions. However, this approach is not suitable for our study because we need to fuse a Gaussian distribution and a Beta distribution, which are heterogeneous distributions, and the discrepancy between a Gaussian and a Beta distribution is not well-defined. Therefore, in this study, we propose a new Likelihood Maximization-based Fusion Mechanism (LMFM) by maximizing the likelihood of generating samples. Formally, we first draw a set of samples from the Gaussian distribution and a set of samples from Beta distribution, denoted as $\{s_l^{\text{Gaus}}\}$ and $\{s_l^{\text{Beta}}\}$, respectively. The fused distribution is then obtained by

$$f_{\text{fused}} = \text{argmax}_f \prod_{s \in \{s_l^{\text{Gaus}}\}} p(s|f) \prod_{s \in \{s_l^{\text{Beta}}\}} p(s|f) \tag{11}$$

The proposed LMFM operates without the need to compute the discrepancy between the two distributions, overcoming a major obstacle in fusing heterogeneous distributions. The distribution $f$ can be any type as long as it maximizes the likelihood. However, it is impractical to enumerate all possible types of distributions and find the best one. Thus, in practice, we can assume the type of distribution $f$ by human domain knowledge, such as assuming $f$ to be a Gaussian or uniform distribution. A new challenge arises: determining the parameters of $f$. A straightforward approach is to optimize the parameters of $f$ directly. However, this approach requires an iterative optimization for each distribution type, making it inefficient and unstable. Since the distributions from machine intelligence and crowd intelligence change continuously during the learning process, this would result in iterative optimization at each update. To address this, we propose a neural network-based encoding process that directly obtains $f$ by encoding the complete statistics of the Gaussian and Beta distributions. We adopt a neural network for encoding because it possesses excellent transformation capabilities. Without loss of generativity, we assume that the fused distribution is Gaussian, with mean ($\mu$) and variance ($\sigma^2$) parameters being unknown. We use a two-layer MLP whose input is the concatenation of the complete statistics. Formally,

$$A = [E_{\text{machine}}, \text{std}_{\text{machine}}, E_{\text{crowd}}, \text{std}_{\text{crowd}}] \tag{12}$$

$$H = \tanh(V_h A + b_h) \tag{13}$$

$$[\mu, \log \sigma] = V_o H + b_o \tag{14}$$

$\text{std}_{\text{machine}}$ and $\text{std}_{\text{crow}}$ are standard deviations of the machine intelligence (a Gaussian distribution) and the crowd intelligence (a Beta distribution), respectively. $V_h$, $b_h$, $V_o$, and $b_o$ are the learnable parameters of the MLP.

Here we encode $\log \sigma$ rather than $\sigma^2$ because $\log \sigma$ can take both positive and negative values, providing greater flexibility in modeling. Once $\mu$ and $\sigma^2$ are determined, the fused distribution (i.e., Gaussian distribution or uniform distribution) is defined. A key advantage of the fused distribution is that it considers the reliability of both crowd intelligence and machine intelligence, making it more suitable for detection tasks. Another advantage of this approach is its ability to accommodate a range of parameter values, allowing us to compute an integrated result over these parameters., i.e.,

$$\hat{y} = \int_y y \cdot f_{\text{fused}}(y|\sigma, \mu)dy \tag{15}$$

where $\hat{y}$ is the predicted veracity of a given piece of news $x$. This is actually the mean value of the distribution. In this way, we can compute the mean value to gauge the veracity of a piece of news.

### 3.4 Learning the Parameters of The Proposed Method

The goal of the first two modules is to provide predictions that enable higher precision in detecting fake news. Therefore, the parameters of these modules are updated by maximizing the likelihood of the ground-truth, denoted as $f(y_i|x_i)$, where $y_i$ is the correct label of news $x_i$. For the third module, its main goal is to obtain a high-quality fused distribution based on the output distributions of the first two modules, hence the likelihood defined in Equation (12) is the goal. Formally, let the parameters of the first two modules be $\boldsymbol{\theta}_1$ and $\boldsymbol{\theta}_2$, where $\boldsymbol{\theta}_1$ refers to the weights ($\boldsymbol{w}$) of the Bayesian deep learning model, and $\boldsymbol{\theta}_2$ encompasses users' reliability $\{c_j\}$. We also denote the parameters of the third module as $\boldsymbol{\theta}_3$, which includes $V_h$, $b_h$, $V_o$, and $b_o$. The optimal $\boldsymbol{\theta}_1$, denoted as $\boldsymbol{\theta}_1^*$, is given by,

$$\boldsymbol{\theta}_1^* = \text{argmax}_{\boldsymbol{\theta}_1} \prod_i f(y_i|x_i, \boldsymbol{\theta}_1) \tag{16}$$

By taking the logarithm, this can be rewritten as,

$$\boldsymbol{\theta}_1^* = \text{argmax}_{\boldsymbol{\theta}_1} \sum_i \log\big(f(y_i|x_i, \boldsymbol{\theta}_1)\big) \tag{17}$$

Since the ground-truth $y_i$ is either 0 or 1, this can be expressed in cross-entropy form as:

$$\boldsymbol{\theta}_1^* = \text{argmin}_{\boldsymbol{\theta}_1} - \sum_i [y_i \log\big(f(y_i = 1|x_i, \boldsymbol{\theta}_1)\big) + (1 - y_i) \log\big(f(y_i = 0|x_i, \boldsymbol{\theta}_1)\big)] \tag{18}$$

However, obtaining the optimal $\boldsymbol{\theta}_1^*$ is challenging due to the presence of local minima. As is common practice in deep learning, we adopt gradient descent methods such as SGD to update $\boldsymbol{\theta}_1$ gradually. Formally, denoting the term we hope to minimize as $L_{\text{machine}} = -\sum_i [y_i \log\big(f(y_i = 1|x_i, \boldsymbol{\theta}_1)\big) + (1 - y_i) \log\big(f(y_i = 0|x_i, \boldsymbol{\theta}_1)\big)]$, then,

$$\boldsymbol{\theta}_1 \leftarrow \text{SGDOptimizer}\big(\boldsymbol{\nabla}_{\boldsymbol{\theta}_1} L_{\text{machine}}, \boldsymbol{\theta}_1\big) \tag{19}$$

Similarly, $\boldsymbol{\theta}_2$ is updated as follows:

$$\boldsymbol{\theta}_2 \leftarrow \text{SGDOptimizer}\big(\boldsymbol{\nabla}_{\boldsymbol{\theta}_2} L_{\text{crowd}}, \boldsymbol{\theta}_2\big) \tag{20}$$

where $L_{\text{crowd}}$ has been defined in Equation (6). For $\boldsymbol{\theta}_3$, the goal is to maximize the likelihood derived from the samples, which is equivalent to minimizing the negative log-likelihood:

$$\boldsymbol{\theta}_3 \leftarrow \text{SGDOptimizer}\left(\boldsymbol{\nabla}_{\boldsymbol{\theta}_3}\left(-\sum_{s\in\{s_l^{\text{Gaus}}\}} \log p(s|f) - \sum_{s\in\{s_l^{\text{Beta}}\}} \log p(s|f)\right), \boldsymbol{\theta}_3\right) \quad (21)$$

## 4. Evaluation

First, we describe the dataset used for the evaluation. Next, we outline the baseline algorithms and the evaluation metrics used for comparison. Then, we detail the experiments we conducted and present the results we obtained.

### 4.1 Dataset Description

We selected a publicly available Weibo rumor dataset to evaluate our model. Sina Weibo is a popular public online social media platform in China. Faced with the huge market demand in China, its daily traffic reaches billions, making it an important platform for information release and dissemination. The Weibo rumor dataset includes 1538 rumors and 1849 non-rumors. They enter the website backend management system through the API interface provided by Weibo to obtain the required news source information and all forwarding/commenting information. The Weibo rumor dataset specifically includes two parts: Weibo original texts with non-rumor and rumor tags, as well as forwarded/commented content with user IDs and posting times. The original texts of the Weibo dataset can be analyzed by machine learning, while the commented contents reflect crowd intelligence.

We further preprocessed the data with the following steps. 1) Removing special characters: As suggested by previous studies (Truong and Tran 2023), special characters such as punctuation marks and emojis that appear in news are removed. 2) Segmenting words: To enable the extraction of Chinese text features, this study utilizes the Jieba segmentation tool (Sun 2012) to process the original Chinese news text, breaking it down into distinct sets of words.3) Removing stop words: We refer to the Baidu Chinese stop word list (Na and Xu 2015) to remove stop words such as "these", "I", "is". 4) Removing null values: Since some news texts become null after removing stop words, we remove these null values. 5) Dividing the dataset: We randomly divided the mixed news data into training, validation, and testing dataset with a ratio of 7:2:1.

We then denoted the attitude of each user's comments. Since some users participated in commenting only a few times, data sparsity makes it hard to accurately measure their reliability.

Therefore, we retained users who have commented on the news at least 5 times. For these users, we invited three experts from university to annotate their stance on the news based on their comments, indicating whether they supported or opposed the news. Specifically, when users believe the news is truthful, they express a supportive attitude in their comments, whereas when users believe the news is fake, they express an opposing attitude. Based on the annotations from the three experts, a majority vote was conducted to determine the user's final stance. Finally, the dataset samples used in this study are presented in Table 1.

**Table 1. Weibo Data Set Statistics**

| Number of News | Number of Users | Number of comments | News length |
|---|---|---|---|
| Total: 1900 | Total: 1702 | Total: 9697 | Average: 31.8 |
| True news: 931 | — | Support attitude:7910 | Longest: 59 |
| Fake news: 969 | — | Opposing attitude:1787 | Shortest: 2 |

## 4.2 Experiments

We evaluate the model with accuracy, precision, recall, and F1. Additionally, we include the area under the receiver operating characteristic curve (AUC). For all five metrics, a higher value indicates better performance.

### 4.2.1 Experiment 1: Comparison with Baselines

As mentioned in the literature review, the existing methods can be divided into three categories: 1) machine intelligence-based, 2) crowd intelligence-based, and 3) machine and crowd intelligence-based methods. Hence, the baselines in the experiment also include the three types.

For the machine intelligence-based ones, the existing methods are mainly based on CNN, LSTM, GRU, or Transformer. Hence, this study includes a CNN-based method (Kim 2014), an LSTM-based method (Bankar and Gupta 2023), a GRU-based method (Aslam et al. 2021), and a BERT-based method (Wu et al. 2023).

For the crowd intelligence-based methods, we compare our method with majority voting (MV) (Penrose 1946), weighted voting (WV), and unsupervised fake news detection framework (UFD) (Yang et al. 2019). The MV method determines the veracity of news by the attitude of the majority users' comments. WV assigns a weight for different users based on the accuracy of user recognition of news, and then determines the veracity of news by the attitude which obtains a larger sum of the weight. UFD creates a probability relationship between the reliability of users commenting news and the authenticity of the news with the Bayesian network, and it infers the authenticity of news by Bayesian learning.

Baselines combining machine and crowd intelligence include SVM (Tran et al. 2020), TCNN-URG (Qian et al. 2018), MFAN (Zheng et al. 2022), HAGNN (Xu et al. 2023), MMVED (Xie et al. 2020), MRHFR (Wu et al. 2023), and Soft Voting (Nielsen 2022). The SVM baseline calculates the proportion of tweets forwarded or commented on and concatenates it with Word2vec-based text representations, using SVM for classification. TCNN-URG employs a Two-Level Convolutional Neural Network (TCNN) to condense word-level information into sentence-level representations, combined with a User Response Generator (URG) that models user interactions for final classification. MFAN uses attention mechanisms to fuse features from news text, images, and propagation graphs, where user comments initialize graph nodes, followed by binary classification. HAGNN applies Graph Neural Networks (GNNs) to integrate text representations with rumor propagation structures, leveraging a Graph Convolutional Network (GCN) for event representations. MMVED employs an MLP to learn distributional features from multimodal inputs, similar to an encoder module. MRHFR calculates the consistency between news and user comments, using it as input for classification. Soft Voting averages prediction probabilities from multiple models, combining machine and crowd intelligence, with a threshold applied to determine the final class.

The comparison results are shown in Table 2, where the bold black text indicates the best results. Our models include two variants: RAHI (Gaussian), which integrates machine intelligence and crowd intelligence as a Gaussian distribution, and RAHI (Uniform), which integrates them as a uniform distribution. We observe that both RAHI (Gaussian) and RAHI (Uniform) achieved good performance. For instance, RAHI (Gaussian) achieved an accuracy of 89.84%, a precision of 89.88%, a recall of 89.91%, an F1 score as 89.84%, and the AUC as 94.23%, outperforming all the existing baselines. Machine-crowd collaborative baselines were among the most competitive methods, but their performance remained inferior to that of our RAHI models. Additionally, Table 2 shows that RAHI (Gaussian) and RAHI (Uniform) have comparable performance, suggesting that treating the fusion distribution as either Gaussian or uniform can yield similar downstream task performance. Therefore, we primarily present the results of RAHI (Gaussian) and refer to it as RAHI for simplicity.

**Table 2. Experimental Results of Comparison with Baselines**

| Methods | | Accuracy | Precision | Recall | F1 | AUC |
|---|---|---|---|---|---|---|
| Machine | CNN | 81.25% | 79.25% | 81.45% | 80.03% | 88.99% |
| | LSTM | 86.72% | 86.75% | 86.78% | 86.72% | 90.20% |
| intelligence | GRU | 82.81% | 82.84% | 82.75% | 82.77% | 89.71% |
| | BERT | 84.38% | 84.41% | 84.31% | 84.34% | 93.91% |
| Crowd intelligence | MV | 58.59% | 70.13% | 57.36% | 50.03% | — |
| | Weighted voting | 62.50% | 78.95% | 61.29% | 55.09% | 58.33% |
| | UFD | 65.63% | 80.00% | 64.52% | 60.00% | 61.58% |
| Machine-crowd collaborative intelligence | SVM | 71.88% | 73.70% | 71.41% | 71.02% | — |
| | TCNN-URG | 82.81% | 84.48% | 79.03% | 81.67% | 86.78% |
| | MFAN | 88.95% | 88.91% | 88.13% | 88.33% | — |
| | HAGNN | 88.20% | 87.10% | 88.40% | 81.60% | — |
| | MMVED | 86.72% | 87.82% | 86.42% | 86.44% | 91.34% |
| | MRHFR | 88.28% | 88.39% | 88.20% | 88.25% | 91.69% |
| | Soft voting | 89.06% | 89.79% | 88.86% | 88.97% | 92.82% |
| Our method | **RAHI (Gaussian)** | **89.84%** | **89.88%** | **89.91%** | **89.84%** | **94.23%** |
| | RAHI (Uniform) | 89.06% | 89.24% | 88.98% | 89.00% | 93.50% |

#### 4.2.2 Experiment 2: Ablation Analysis

**1) Advantage of Combining Both Types of Intelligence**

We compare our method with the variants where only machine intelligence or crowd intelligence is exploited.

**Table 3. Ablation Analysis of Crowd Intelligence and Machine Intelligence**

| Methods | Accuracy | Precision | Recall | F1 | AUC |
|---|---|---|---|---|---|
| Machine intelligence-based | 87.50% | 87.50% | 87.54% | 87.50% | 94.01% |
| Crowd intelligence-based | 74.22% | 76.95% | 73.61% | 73.04% | 79.61% |
| Machine-crowd collaboration-based | **89.84%** | **89.88%** | **89.91%** | **89.84%** | **94.23%** |

The experimental results are shown in Table 3. There are two observations. First, the results show that performance decreased when machine intelligence was removed. Specifically, the crowd intelligence-based method achieved an accuracy of 74.22%, a precision of 76.95%, a recall of 73.61%, an F1 score of 73.04%, and an AUC of 79.61%, which were significantly lower than the methods where machine intelligence is involved. Second, after removing crowd intelligence, the performance of our model was also reduced. For instance, the detector achieved an accuracy of 87.50%, precision of 87.50%, recall of 87.54%, F1 score of 87.50%, and AUC of 94.01%, which were significantly lower than the method where crowd intelligence was involved, whose corresponding metric values were 89.84%, 89.88%, 89.91%, 89.84%, and 94.23%. These two observations further demonstrate the advantage of combining machine intelligence and crowd intelligence for fake news detection.

**2) Advantage of Considering The Reliability of Machine Intelligence**

We show the advantage of considering reliability by comparing the model that considers (i.e., the Bayesian model with reliability) and that without. In particular, the model without considering reliability is a classic BERT, while the model that considers reliability is a BERT equipped with Bayesian learning. The comparison results are in Table 4.

**Table 4. Abalation Analysis of Considering Reliability of Machine Intelligence**

| Methods | Accuracy | Precision | Recall | F1 | AUC |
|---|---|---|---|---|---|
| Without considering reliability | 84.38% | 84.41% | 84.31% | 84.34% | 93.91% |
| Considering reliability | **87.50%** | **87.50%** | **87.54%** | **87.50%** | **94.01%** |

The experimental results show that when considering reliability, the machine intelligence achieved an increase in accuracy by 3.12%, precision by 3.09%, recall by 3.23%, and F1 score by 3.16%. This demonstrates the effectiveness of incorporating reliability into the machine intelligence module for fake news detection tasks.

**3) Advantage of Adjusting The Reliability of Crowd Intelligence**

In the crowd intelligence module, we propose calculating the reliability of each user based on the IRT statistical theory and then updating it with the supervision-guided user reliability adjustment. To examine the advantage of adjustment, we removed this module and tested the model's performance. The results are shown in Table 5. As shown in the table, the model without adjusting the weights achieved only an accuracy of 85.16%, a recall of 85.41%, an F1 score of 85.11%, and an AUC value of 89.37%, proving the effectiveness of the supervision-guided user reliability adjustment method.

**Table 5. Ablation Analysis of Supervision-Guided User Reliability Adjustment**

| Methods | Accuracy | Precision | Recall | F1 | AUC |
|---|---|---|---|---|---|
| Without adjustment | 85.16% | 86.09% | 85.41% | 85.11% | 89.37% |
| With adjustment | **89.84%** | **89.88%** | **89.91%** | **89.84%** | **94.23%** |

**4.3 Experiment on Dynamic Fake News Detection**

In the early stages of news publication, user responses are sparse, and machine intelligence plays a dominant role. Initially, when no user responses are available, machine intelligence analyzes the news's veracity based solely on its content. Over time, as user responses accumulate, crowd intelligence becomes increasingly significant. This section evaluates the accuracy of fake news detection as user responses grow. In the experiment, the publication time was set as 1 minute, and model performance was assessed at intervals: 1 minute, 2 minutes, 3 minutes, 4 minutes, 5 minutes, 10 minutes, 20 minutes, 30 minutes, 1 hour, 1.5 hours, 2 hours, 3 hours, 4 hours, 5 hours, 12 hours, 24 hours, 2 days, 3 days, 4 days, and 7 days. Results are shown in Figure 2.

**Figure 2. The Accuracy in Detecting Fake News Over Time**

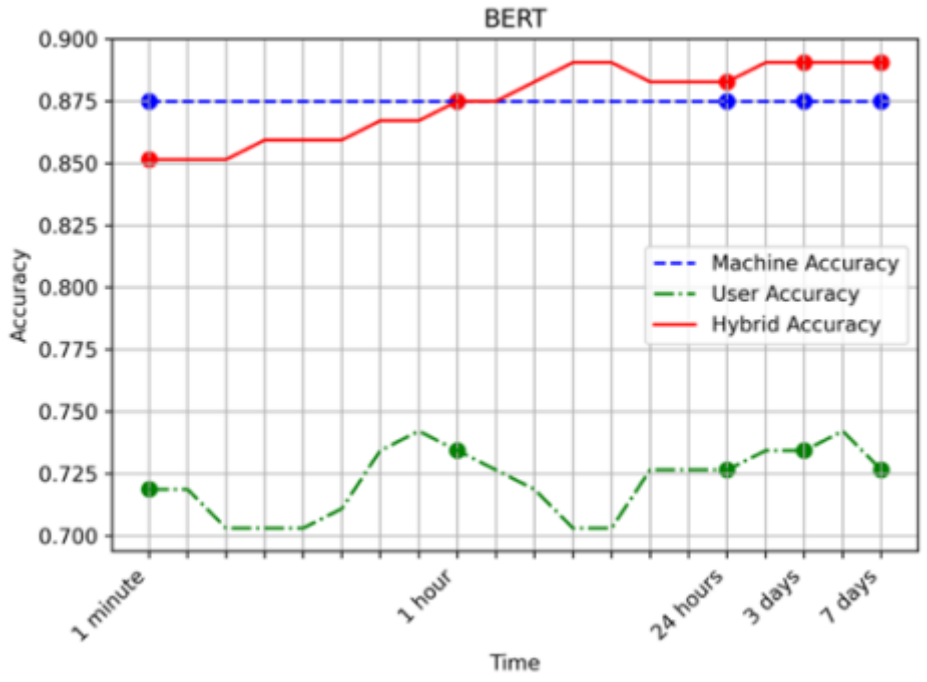

During the experiment, user-generated comments before different time points were used as inputs for the crowd intelligence module. In the beginning, due to the small number of user comments, the predicted results vary significantly with different levels of user participation. Over time, the user's prediction results steadily increase. At the 24-hour node, the prediction accuracy of crowd intelligence remained stable and after that the accuracy increased gradually. In the machine intelligence module, the news content did not change over time, so the prediction results of each machine intelligence remained unchanged during the dynamic detection of fake news. As the time interval increases, the number of user comments increased, the effect of crowd intelligence gradually stabilized, and the effect of machine-crowd collaboration also gradually improved. As shown in Figure 2, the machine-crowd hybrid intelligence fake news detection method improved constantly over time, and demonstrated significant advantages over machine intelligence at a time threshold of 1 hour. We observe that initially, the hybrid intelligence method underperformed the machine intelligence-based method due to the sparsity of crowd responses. However, in practice, when only a few crowd responses are available, it watypically indicates that the news has been noticed by fewer users, and thus its societal impact is limited. As a result, platform managers are less likely to prioritize such news. In contrast, as the news begins to gain traction—evidenced by an increase in user responses—our hybrid approach catched up and eventually outperformed the machine intelligence-based method. This makes it more effective for detecting fake news in scenarios where user engagement grows dynamically.

**4.4 Case Study**

In this section, we used several examples to demonstrate how our model works. We used BERT as the machine distribution discrimination method, combined with the output results of user results, to provide the fused distribution results, as shown in Figure 3. The distribution of the fused model (represented by the red curve) exhibited a strong alignment with the true label 1, while the machine model (blue curve) demonstrated a bias towards label 0. In contrast, the crowd model (green curve) showed considerable variability around label 1. When these distributions were integrated, the resulting combined distribution shifted towards label 1, indicating that incorporating both machine and crowd models enhanced the overall decision-making accuracy. As illustrated in the right figure, when discrepancies arose between the machine and crowd models, the fusion process effectively mitigated these differences, yielding a more accurate prediction that aligned with the true label. This fusion approach significantly improved both prediction accuracy and reliability, leveraging the strengths of both the machine and crowd models while minimizing their individual weaknesses. Overall, Figure 3 demonstrates that the fusion distribution method substantially enhances prediction accuracy compared to using the BERT model or the crowd distribution alone. By combining these two sources, the approach not only mitigates the limitations of each individual method but also offers a robust solution for addressing complex fake news data. This method enables the model to generate predictions that are consistently more aligned with real-world scenarios, underscoring the efficacy of integrating machine intelligence with crowd intelligence to bolster predictive performance.

**Figure 3. Examples of Case Studies**

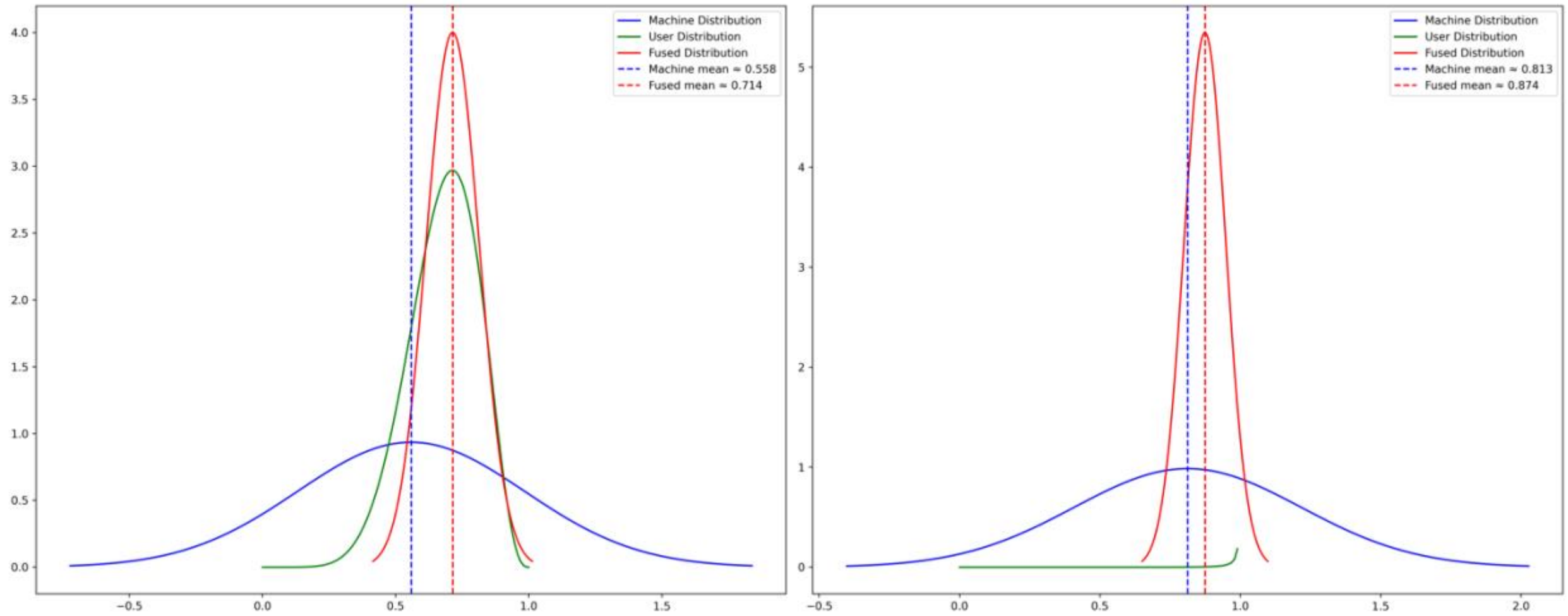

## 5. Conclusion

Social media platforms empower users to share news from their own perspectives and experiences. However, the dissemination of fake news on social media platforms underscores the urgent need for advanced detection methods. Three types of methods have been proposed for the detection: machine intelligence-based methods, crowd intelligence-based methods, and hybrid intelligence-based methods. Among them, hybrid intelligence-based methods achieve the best performance. However, hybrid intelligence-based methods fail to consider reliability in detection, which motivates us to propose a novel Reliability Aware Hybrid Intelligence (RAHI)-based method for fake news detection. This method comprises three integrated modules. The first module employs a Bayesian deep learning model to capture the inherent reliability of machine intelligence. The second module is the Item Response Theory (IRT)-based user response aggregation to account for the reliability of crowd intelligence. The third module is a new distribution fusion mechanism (i.e., likelihood maximization-based fusion mechanism, LMFM), which takes the assessment distributions derived from both machine and crowd intelligence as input, and outputs a fused distribution that offers predictions along with the associated reliability in fake news detection. Based on four sets of experiments, we empirically demonstrate the advantages of our method. This study contributes to the research field by proposing a novel RAHI-based method, a novel user reliability evaluation method, and a novel distribution fusion mechanism.

This study has practical implications for three key stakeholders: internet users, online platform managers, and the government. First, internet users increasingly rely on social media platforms for news about global events. However, the rise of fake news on these platforms undermines source credibility and can mislead users into harmful behaviors, such as social panic. The model proposed in this study helps users identify fake news promptly, protecting them from deception. Second, the prevalence of fake news poses challenges for online platform managers, hindering their efforts to create trustworthy environments. Our study presents an effective model for fake news detection, enabling managers to identify misleading content and enhance platform credibility. Additionally, our method allows for a more reliable assessment of news veracity, reducing the risk of making mistakes. Third, fake news also impacts government operations, making it crucial to address for effective public management. By dynamically detecting fake news, our model can assist the government in monitoring it. Awareness of the status of fake news allows governments to take appropriate measures to counter it, which is vital for fostering a trustworthy society and ensuring effective governance.

Same as other studies, this study also has several limitations and corresponding promsing directions. First, the impact of news detection difficulty on the reliability of crowd intelligence may be influenced by various external factors, such as social platform mechanisms and event types.

Future research can explore these influences in greater depth by utilizing more comprehensive data from social media. Second, the dataset used in this study is based on real data from the Weibo platform. It does not differentiate between AI-generated and human-written fake news. Investigating these distinctions and developing methods to identify AI-generated fake news are critical areas for future research. Third, this study only focuses on textual fake news. However, short-form videos are increasingly prominent in spreading fake news. Hence, future efforts can focus on this type of data in detecting fake news.

**Appendix A**

**Table A.1 Selected Machine Intelligence-based Studies**

| Year | Author | Data Type | Model/Method |
|---|---|---|---|
| 2024 | Peng et al. | News Text, Images | CSFND: The proposed model integrates local context features with global semantic features through unsupervised context learning to enhance multimodal fake news detection. |
| 2023 | Chen et al. | News Text | LSTM, GRU, BiLSTM: The proposed model learns the textual characteristics of each type of true and misinformation for subsequent true/false prediction. |
| 2023 | Zhang et al | News Text | Conv-FFD： Considering the fact that the news are generally short texts and can be remarkably featured by some keywords, convolution-based neural computing framework is adopted to extract feature representation for news texts. |
| 2022 | Mehta et al. | News Text; Network | GCN: The proposed method frames fake news detection as reasoning over the relationships in a graph, using inference operators to uncover unobserved interactions between sources, articles, and users' engagement patterns. |
| 2021 | Wu et al. | News Text, Images | MCAN: The proposed method learns multimodal fusion representations by extracting features from text and image across different domains, fusing them using a deep co-attention model, and then using the fused representation for fake news detection. |
| 2021 | Lee | News Text | FENDI: The proposed method extracts deceptive and semantic cues from text using a combination of deep learning and multiple NLP techniques. |
| 2021 | Yuan et al. | News Text; Network | DAGA-NN: The proposed model uses domain-adversarial and graph-attention techniques to achieve accurate cross-domain fake news detection with limited or no sample data. |
| 2020 | Bian et al. | News Text; Network | Bi-GCN: A GCN with a top-down directed graph of rumor spreading to learn the patterns of rumor propagation; and a GCN with an opposite directed graph of rumor diffusion to capture the structures of rumor dispersion. |
| 2020 | Przybyla | News Text | BiLSTMAvg: The proposed model apply a two-stage approach for selecting relevant features: first preliminary filtering, then building a regularized classifier. |
| 2020 | Kaur et al. | News Text | Ensemble Model: ML classifiers are combined based on their false prediction ratio. |
| 2020 | Ahmad et al. | News Text | Ensemble Model: The proposed model train a |
| | | | combination of different machine learning algorithms using various ensemble methods. |

**Table A.2 Selected Crowd Intelligence-based Studies**

| Year | Author | User Responses | Aggregation Method |
|---|---|---|---|
| 2024 | t'Serstevens et al. | Interactions | Hierarchical Bayesian: The proposed model learn from the crowd-workers, and generate predicted veracity scores for a series of personae, taking into account the attributes of the persona, the tweet and their interactions. |
| 2023 | Truong and Tran | Comments; Likes; Shares | Voting Ensemble Classifier: The proposed method extracts the features from a Twitter dataset and then a voting ensemble classifier comprising three classifiers and Softmax is used to classify news into two categories which are fake and real news. |
| 2023 | Wu et al. | Comments | Selected Mechanism: The proposed method design selected mechanism to extract top-K comments, which calculates the difference between each comment and other comments in an automated manner. |
| 2022 | Benjamin and Raghu | Comments | Random Forest: Human reactions to social bot messages are used to augment existing social bot detection capabilities. |
| 2021 | Souza Freire et al. | Implicit Opinions | HCS: An approach based on crowd signals that considers implicit user opinions instead of the explicit ones |
| 2019 | Yang et al. | Social Engagements | Collapsed Gibbs Sampling: Gibbs sampling approach is a widely-used MCMC method to approximate a multivariate distribution when direct sampling is intractable. |
| 2019 | Pennycook and Rand | Questionnaire | Linear Regressions: This study uses linear regressions predicting trust, with the rating as the unit of observation and robust SEs clustered on participant. |

**Table A.3 Selected Hybrid Intelligence-based Methods**

| Year | Author | Data Type | Fusion Method | Model/Method |
|---|---|---|---|---|
| 2024 | Wang et al. | Claims; Report | Interactive Concatenation | L-Defense: The proposed method proposes an evidence extraction module to split the wisdom of crowds into two competing parties and respectively detect salient evidences. |
| 2023 | Wu et al. | News Text; Images; | Feature Concatenation | MRHFR: The proposed method summarizes three basic cognitive reading |
| | | Comment Text | | habits and put forward cognition-aware fusion layer to learn the dependencies between multimodal features of news. |
| 2023 | Luvembe et al. | Publisher Emotions | Feature Concatenation | BiGRU, AGWu-RF: The study proposes a Deep Normalized Attention-based mechanism for enriched extraction of dual emotion features and an Adaptive Genetic Weight Update-Random Forest for classification. |
| 2023 | Xu et al. | News Text; Comment Text | Feature Concatenation | HAGNN: The proposed method applies a Graph Convolutional Network (GCN) with a graph of rumor propagation to learn the text-granularity representations with the spreading of events. |
| 2022 | Wei et al. | News Text; Report; Comment Text | Bayesian Network | Customized Bayesian Network: The proposed method extracts relevant human and machine judgments from data sources including news features and scalable crowd intelligence. The extracted information is then aggregated by an unsupervised Bayesian aggregation model. |
| 2022 | Mehta et al. | News Text; Comment Text | Feature Concatenation | GCN: The proposed method captures the interaction between social information and news content using a heterogeneous graph and use a Relational Graph Convolutional Network, to create vectorized node representations for factuality prediction. |
| 2021 | Dou et al. | News Text; Propagation Graph | Feature Concatenation | UPFD: The proposed method simultaneously captures various signals from user preferences by joint content and graph modeling. |
| 2021 | Albahar | News Text; Comment Text | Feature Concatenation | RNN, SVM: A hybrid model comprising a recurrent neural network (RNN) and support vector machine (SVM) is incorporated to detect real and fake news. |
| 2019 | Yang et al. | Like or Dislike | Bayesian Network | Customized Bayesian Network: This paper leverage a Bayesian network model to capture the conditional dependencies among the truths of news, the users' opinions, and the users' credibility. |